\documentclass[conference]{IEEEtran}
\IEEEoverridecommandlockouts

\usepackage{cite}
\usepackage{amsmath,amssymb,amsfonts}
\usepackage{algorithmic}
\usepackage{graphicx}
\usepackage{textcomp}
\usepackage{xcolor}
\usepackage{multirow}
\usepackage{booktabs}
\usepackage{subcaption}
\usepackage{url}
\usepackage{tabularx}
\newcolumntype{Y}{>{\centering\arraybackslash}X}
\usepackage[switch]{lineno}

\def\BibTeX{{\rm B\kern-.05em{\sc i\kern-.025em b}\kern-.08em
    T\kern-.1667em\lower.7ex\hbox{E}\kern-.125emX}}
\begin{document}
\bstctlcite{IEEEexample:BSTcontrol}
\title{Variational Contrastive Learning \\ for Skeleton-based Action Recognition
\thanks{ ACKNOWLEDGMENT: This work was partly funded by the AdeptHuman HIP-2024-016 IP PARIS grant and France 2030 ICON project. }
}

\author{\IEEEauthorblockN{Dang Dinh NGUYEN}
\IEEEauthorblockA{\textit{SAMOVAR, Télécom SudParis} \\
\textit{Institut Polytechnique de Paris}\\
dang-dinh.nguyen@telecom-sudparis.eu}
\and
\IEEEauthorblockN{ Decky ASPANDI-LATIF}
\IEEEauthorblockA{\textit{SAMOVAR, Télécom SudParis} \\
\textit{Institut Polytechnique de Paris}\\
decky.aspandi\_latif@telecom-sudparis.eu}
\and
\IEEEauthorblockN{ Titus ZAHARIA}
\IEEEauthorblockA{\textit{SAMOVAR, Télécom SudParis} \\
\textit{Institut Polytechnique de Paris}\\
titus.zaharia@telecom-sudparis.eu}
}

\maketitle

\begin{abstract}

In recent years, self-supervised representation learning for skeleton-based action recognition has advanced significantly via contrastive learning. However, most of contrastive paradigms are inherently discriminative and often struggle to capture the variability and uncertainty intrinsic to human motion. To address this limitation, we introduce a variational contrastive learning framework that integrates probabilistic latent modeling into a self-supervised contrastive pipeline. By mapping skeleton sequences to continuous probability distributions rather than static point-vectors, this formulation enforces a well-structured, regularized latent space. This enhanced representation space preserves distinct class boundaries and demonstrates high robustness against data scarcity in data-contained environments. Extensive experiments on three benchmark datasets show that our method consistently outperforms state-of-the-art approaches, particularly establishing a commanding advantage in low-label regimes ($1\%$ and $10\%$). Furthermore, qualitative analyses confirm that our probabilistic feature space exhibits superior semantic meaningfulness, yielding better-separated class clusters and more precise joint-level spatial attention maps compared to deterministic baselines.

\end{abstract}



\section{Introduction}
Human action recognition remains a pivotal challenge in computer vision due to significant intra-class variability, complex spatio-temporal dynamics, and subtle nuances between actions. Despite these complexities, robust recognition is essential for applications from surveillance to human-computer interaction \cite{Ren2020ASO, s19051005}. Recently, 3D skeleton keypoints have emerged as an effective modality, offering a compact, semantically rich motion encoding that minimizes background pixel dependencies and simplifies cross-view spatial mapping \cite{zhang2025selfsupervisedskeletonbasedactionrepresentation}.


Although supervised skeleton-based methods achieve high recognition accuracy, they rely heavily on large labeled datasets, which restricts their scalability in real-world scenarios where annotated samples are scarce \cite{zheng2018unsupervised, Lin_2020}. Self-supervised learning (SSL) offers a promising alternative, with contrastive learning emerging as a dominant paradigm. By matching different augmented views of the same sequence without manual labels, contrastive methods force the model to map identical action classes into distinct, well-separated feature clusters. However, this discriminative nature prevents the models from capturing the inherent variability and uncertainty of human motion - for instance, subtle differences in how the same action is performed - limiting its ability to represent the full range of plausible movements.




In this work, we introduce a unified variational contrastive learning framework that extends deterministic self-supervised paradigms by mapping skeleton sequences to continuous probability distributions.  The proposed formulation enforces a continuous, regularized latent structure capable of capturing  stylistic variability and spatial uncertainties inherent to human motion. By mathematically synthesizing the discriminative capabilities of contrastive learning with the structured generative traits of variational inference, our framework establishes a semantically grounded feature manifold that effectively disentangles complex spatio-temporal dynamics from skeleton configurations. Our main contributions are  as follows:


\begin{enumerate}
    \item We propose a unified variational contrastive learning framework that mathematically unifies self-supervised contrastive paradigms with probabilistic variational modeling, providing a structurally elegant architecture specifically tailored for skeleton-based representations.

    \item We conduct extensive empirical evaluations across standard skeleton-based action recognition benchmarks under diverse testing protocols, demonstrating robust, competitive performance on downstream classification tasks—particularly establishing a strong advantage in low-label semi-supervised regimes ($1\%$ and $10\%$).

    \item We provide detailed qualitative analyses and latent space visualizations demonstrating that our probabilistic representations preserve superior semantic consistency and enforce more precise joint-level spatial attention maps compared to deterministic baselines.
\end{enumerate}

The remainder of this paper is organized as follows: Section 2 reviews related work on self-supervised skeleton-based action recognition, and variational inference. Section 3 introduces the proposed methodology. Section 4 details the experimental setup and presents the results. Finally, section 5 concludes our work and opens perspectives of future research.

\section{Related Work}


\subsection{Self-supervised skeleton-based human action recognition}

Skeleton-based action recognition has evolved from rigid handcrafted descriptors \cite{6247813, 6909476} to deep architectures like RNNs \cite{7298714, 8322061} and CNNs \cite{Ke_2017_CVPR, 8026285} that automate feature extraction. However, such early  models ignore the non-Euclidean topology of the human body. Graph Convolutional Networks such as ST-GCN \cite{yan2018spatialtemporalgraphconvolutional} resolved this by modeling skeletons as spatio-temporal graphs. Yet, the reliance of these architectures on expensive annotations has accelerated the modern shift toward Self-Supervised Learning (SSL).

Existing SSL approaches broadly fall into three categories. A first one employs generative and context-based pretext tasks—including LongT GAN \cite{zheng2018unsupervised}, P\&C \cite{su2020predict}, TS-Colorization \cite{yang2021skeletoncloudcolorizationunsupervised}, SeBiReNet \cite{nie2021unsupervised3dhumanpose}, and GL-Transformer \cite{kim2022globallocalmotiontransformerunsupervised}. By utilizing surrogate tasks such as input reconstruction, future prediction, or colorization, such methods encourage the model to capture key spatio-temporal dependencies. A second family concerns the contrastive learning methods, such as AS-CAL \cite{rao2021augmentedskeletonbasedcontrastive}, ISC \cite{thoker2021skeletoncontrastive3dactionrepresentation}, and CrosSCLR \cite{li20213dhumanactionrepresentation}, which enforce consistency between differently augmented views of the same sequence. Finally, hybrid approaches \cite{Lin_2020,xu2020prototypicalcontrastreverseprediction} combine multiple  tasks to increase the degree of generalization of the  representations.

In this work, we focus on the self-supervised contrastive paradigm, given its proven efficacy in establishing discriminative latent spaces for skeleton-based representations.

\subsection{Variational inference}
While SSL has advanced skeleton-based action recognition, existing methods tend to focus on discriminative modeling and often overlook the probabilistic view of the uncertainty and variability aspects that naturally occur in human motion. In the broader machine learning literature, studies combined variational inference with generative \cite{Gatopoulos_2021} and recently, contrastive SSL \cite{Nakamura_2023_ICCV,yavuz2025variational,10582001}. They demonstrate that probabilistic objectives can complement this learning paradigm. Human motion is characterized by significant intra-class variability - encompassing fluctuations in style, speed, and viewpoint - and inter-class ambiguity stemming from morphologically similar actions. Representing such variations under a probabilistic framework facilitates the acquisition of richer, semantically grounded representations \cite{wang2022understandingcontrastiverepresentationlearning, jeong2025probabilisticvariationalcontrastivelearning}. Some first studies have explored the intersection of variational and contrastive paradigms, but their objectives often diverge from the pursuit of a structured motion  space. For instance, \cite{wu2025frequencysemanticenhancedvariationalautoencoder} prioritizes cross-modal alignment for zero-shot tasks, while Skeleton-logoCLR \cite{10551297} utilizes this combination as a consistency tool to synchronize similarity distributions across views. In contrast, our work shifts from deterministic alignment to probabilistic modeling, employing a variational regularizer to explicitly account for motion uncertainty and variability - a perspective that remains under-explored in the literature.

\section{Proposed methodology}
\subsection{Theoretical background}
\label{sec:subsec:theo}





To address the  limitations of deterministic contrastive learning, we establish a connection between its categorical discrimination 
and the probabilistic framework of variational inference.

\textbf{\emph{The Evidence Lower Bound} (ELBO)}: Consider a latent-variable model where $x$ is the observed skeleton sequence and $z$ is its latent representation. The log-likelihood $\log p(x)$ can be lower-bounded by the ELBO:
\begin{equation}
\begin{aligned}
    \log p(x) 
    &\geq \mathcal{L}_{\text{ELBO}}(\phi) \\
    &= \mathbb{E}_{q_\phi(z \mid x)} \big[ \log p(x \mid z) \big] - D_{\text{KL}}\big( q_\phi(z \mid x) \,\|\, p(z) \big)
\end{aligned}
\end{equation}
where $q_\phi(z \mid x)$ is the variational posterior (encoder) and $p(z) \thicksim \mathcal{N}(0,I_d)$ is the prior. The first term is the reconstruction likelihood, while the second regularizes the latent space.

\textbf{Contrastive reconstruction as surrogate likelihood.}
In high-dimensional skeleton spaces, pixel-wise or joint-wise reconstruction $p(x \mid z)$ is often noisy and fails to capture high-level semantics. Following \cite{wang2022understandingcontrastiverepresentationlearning, jeong2025probabilisticvariationalcontrastivelearning}, we replace the raw reconstruction with a categorical surrogate. Instead of reconstructing $x$ directly, the model predicts which embedding $z'$ (from a positive augmented view) corresponds to $z$ among a set of negative candidates. Formally, one can define:
\begin{equation}
    p(z' \mid z) = \frac{p(z,z')}{\int p(z,z'')\,dz''} 
\end{equation}

Since $p(z, z')$ is unknown, it is then approximated using a similarity-based exponential scoring function. Normalizing over a set of negative samples $\{z'_j\}_{j=1}^{J}$ yields
\begin{equation}
    p(z' \mid z) \;\approx\;
    \frac{\exp\left(\mathrm{sim}(z, z')\right)}{\sum_{j=1}^{J} \exp\left(\mathrm{sim}(z, z'_j)\right)}.
\end{equation}

which corresponds exactly to the Softmax-normalized similarity used in the InfoNCE loss \cite{oord2019representationlearningcontrastivepredictive}. 

\textbf{Unified Variational Contrastive Objective}
By substituting the expected log-likelihood with the negative InfoNCE loss, we derive the Variational Contrastive (VCL) objective:
\begin{equation}
\ell_{\text{VCL}}(x) = \ell_{\text{InfoNCE}}(x) + 
D_{\text{KL}}\!\big(q_\phi(z \mid x)\,\|\,p(z)\big)
\label{eq:vcl}
\end{equation}
Here, InfoNCE acts as a decoder-free surrogate for reconstruction, while the KL divergence enforces structural consistency on the probabilistic manifold. This casts contrastive learning as a special case of variational inference, bridging discriminative SSL with probabilistic latent-variable modeling. 



\subsection{Overall method}

\begin{figure*}[!hbt] 
    \centering
    \includegraphics[width=\textwidth]{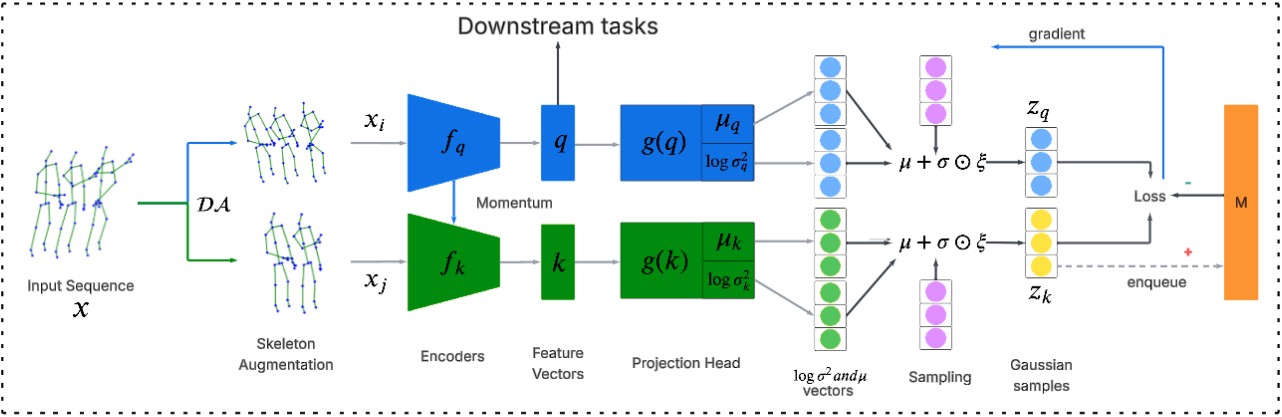}
    \caption{Overview of the proposed framework which includes variational projection heads that enable Gaussian sampling for contrastive pretext training.}
    \label{fig:architecture}
\vspace{-3mm}
\end{figure*}

Fig.~\ref{fig:architecture} illustrates the proposed framework. Our method integrates variational latent modeling into SkeletonCLR \cite{li20213dhumanactionrepresentation}, a foundational MoCo-v2 baseline \cite{chen2020improved} chosen for its  minimalist design. This simple architecture isolates our evaluation, ensuring that all performance gains are driven strictly by our probabilistic feature formatting rather than backbone engineering.

Let us consider a mini-batch of skeleton sequences:
\[
 x = \{ x_1, x_2, \dots x_n \} \in \mathbb{R}^{C \times T \times N}
\]
where $C$ denotes the feature dimension (e.g., joint coordinates), $T$ the number of frames, and $N$ the number of joints per frame. During pretext training, we obtain the skeleton sequence $x$ and generate two different sequences $x_i$ and $x_j$ using a random data augmentation ($DA$), which consists of a shear transformation \cite{rao2021augmentedskeletonbasedcontrastive} with amplitude $\beta = 0.5$ and temporal crop \cite{Shorten2019ASO} with padding ratio $\gamma = 6$. Each sequence is then processed by a separate encoder: the query encoder producing $q$, and the key encoder producing $k$. Instead of mapping encoder outputs directly to a deterministic embedding space, we propose to introduce a Gaussian sampling head $g(\cdot)$ to estimate the distribution parameters:
\[
(\mu_q, \log \sigma^2_q) = g(q), \quad (\mu_k, \log \sigma^2_k) = g(k)
\]

We then apply the reparameterization trick \cite{kingma2013auto} to draw latent samples $z_q$, $z_k$, forming a positive pair for the contrastive loss:
\[
z_q = \mu_q + \sigma_q \xi , \quad z_k = \mu_k + \sigma_k \xi, \quad \xi \sim \mathcal{N}(0,I)
\]


To maintain the consistency of key representations, a momentum update strategy is introduced. The parameters of the key encoder are initialized by copying those of the query encoder. During training, the query encoder is updated via back-propagation, whereas the key encoder is updated as an exponential moving average of the query encoder. Let $\theta_q$ and $\theta_k$ denote the parameters of the query and key encoders, respectively. The update rule for the key encoder is given by: $\theta_k \leftarrow \varepsilon \theta_k + (1 - \varepsilon) \theta_q$, where $\varepsilon \in [0,1]$ is a momentum coefficient controlling the update rate of the key encoder.

For the downstream task of action classification, the feature representations produced by the query encoder are subsequently fed into a fully connected classification layer.

Following the multi-stream paradigm \cite{2sagcn2019cvpr}, we train separate models on joint (3D coordinates), bone (vectors between connected joints), and motion (temporal differences between frames) stream. The final prediction is obtained via weighted fusion with weights [0.6,0.6,0.4], consistent with prior multi-stream GCN-based approaches \cite{Shi_2020,2sagcn2019cvpr}. The implementation of our method is publicly available in our repository
\footnote{\url{https://github.com/Dang-Dinh-NGUYEN/graph-based_action-recognition}}.


\subsection{Objective function}

During pretext training, we minimize the unified Variational Contrastive (VCL) loss introduced in (\ref{sec:subsec:theo}, Eq.~\ref{eq:vcl}). Specifically, the InfoNCE term is computed using the probabilistic samples $z_q$ and  $z_k$ from the query and the key encoders:
\begin{equation}
\begin{aligned}
\ell_{InfoNCE} = & - \log 
\frac{\exp\left(z_q \cdot z_k / \tau \right)}
{\exp\left(z_q \cdot z_k / \tau \right) + \sum_{z_{k}^-} \exp\left(z_q \cdot z_{k}^- / \tau \right)}
\end{aligned}
\end{equation}
where $\tau$ is the temperature parameter (set to $0.07$ in our experiments), $z_{k}^-$ denotes negative samples from the queue, and the similarity measure $z_q \cdot z_k$ is the dot product (equivalent to cosine similarity when embeddings are $\ell_2$-normalized).


To regularize the latent space, we impose a KL divergence towards a Gaussian distribution \cite{odaibo2019tutorial}:
\begin{equation}
\begin{aligned}
    \ell_{z}^{norm} = D_{KL}(q_{\theta}(z|x)\,\|\,\mathcal{N}(0,1))
    = -\frac{1}{2}\big[1 + \log(\sigma^2) - \sigma^2 - \mu^2\big]
\end{aligned}
\end{equation}
where $\log\sigma^2$ and $\mu$ are the outputs of the variational encoder for each latent sample. Finally, the overall pretext training loss is defined as: 
\begin{equation}
    \ell^{total} = \ell_{InfoNCE} + \beta_{kl}(
    \ell_{z_q}^{norm} + \ell_{z_k}^{norm}
    )
\end{equation}

The $\beta_{kl}$ adjust the importance of KL term, ensuring the model is fully cognizant of the probabilistic latent structure. To mitigate posterior collapse, we implement an 
annealing schedule \cite{bowman2016generatingsentencescontinuousspace}. During the initial 50 epochs, $\beta_{kl}$ is linearly increased from $0$ to 1. This "warm-up" phase prioritizes the optimization of the contrastive InfoNCE objective, allowing the encoder to first establish a stable representation before the variational constraints are fully enforced.

For downstream action classification, the sampling head is discarded, and the query encoder is optimized using a standard cross-entropy loss.




\section{Experimental results}


\subsection{Datasets}


\textbf{NTU RGB+D 60 \& 120}: NTU-60 \cite{shahroudy2016ntu} consists of 56,880 sequences across 60 classes, evaluated via Cross-Subject (xsub) and Cross-View (xview) protocols. NTU-120 \cite{liu2020ntu} extends this to 114,480 sequences and 120 classes, utilizing xsub and Cross-Setup (xset) protocols. These benchmarks are standard for assessing robustness across different actors, viewpoints, and environmental configurations.

\textbf{PKU-MMD}: This dataset \cite{10.1145/3132734.3132739} includes 20,000 samples and 52 action classes. It is divided into Part I and a highly challenging Part II, the latter featuring abrupt motion variations and severe view occlusions. Following standard practice, we adopt the xsub protocol for both subsets.

The datasets are partitioned into independent training and testing sets according to their official recommendation.

\subsection{Experiment settings}
\label{sec:exp}



Following \cite{2sagcn2019cvpr}, all skeleton sequences are normalized to a fixed length of 50 frames via linear interpolation. We employ ST-GCN \cite{yan2018spatialtemporalgraphconvolutional} as the primary encoder backbone, reducing each layer's channel capacity to one-fourth to reduce computational burden. While our variational framework is backbone-agnostic and compatible with alternative architectures like AAGCN \cite{Shi_2020}, we retain ST-GCN due to its superior  performance.

The models are implemented in PyTorch and trained from scratch using randomly initialized weights on two NVIDIA RTX 4090 GPUs. During self-supervised pre-training, the model is trained for $300$ epochs strictly on the unlabeled training split. Parameters are optimized via AdamW \cite{loshchilov2019decoupledweightdecayregularization} with a weight decay of $1 \times 10^{-4}$ and an initial learning rate of $0.001$, which is decayed by a factor of $10$ at epoch $250$. The contrastive memory bank size is maintained at $30\text{K}$. Three evaluation protocols are used in our experiment: 
\begin{itemize}
    \item \textbf{Linear Evaluation.} To evaluate the quality of the learned representations, we follow a standard linear evaluation protocol~\cite{rao2021augmentedskeletonbasedcontrastive}. After pre-training, the encoder is frozen, and a linear classifier is trained on top of the fixed embeddings using all available labeled data for 100 epochs. The learning rate of 0.03 is used during training, which is decreased by a factor of 10 at epoch 80.
    \item \textbf{Semi-supervised Evaluation.} Following prior configurations~\cite{thoker2021skeletoncontrastive3dactionrepresentation, yang2021skeletoncloudcolorizationunsupervised}, the pre-trained encoder and a random classification head are fine-tuned end-to-end on limited subsets ($1\%$ and $10\%$) of the training split. To ensure uniform supervision, a category-balanced sampling strategy is employed to select an identical number of samples per action class. All optimization settings, hyperparameters, and augmentations are identical to the linear evaluation protocol. We retain solely the NTU-60 and PKU-MMD datasets to ensure a direct baseline comparison.
    

    \item  \textbf{Fine-tuning.}
    In this fully supervised setting \cite{thoker2021skeletoncontrastive3dactionrepresentation}, the entire pretrained encoder is unfrozen and trained end-to-end with the classifier head using all labeled data. 
\end{itemize}


\subsection{Ablation study}

\textbf{Effectiveness of Variational Contrastive SSL}

\begin{table}[!ht]
\centering
\caption{Linear and Semi-supervised (1\%, 10\%) results on NTU and PKU-MMD. Values in parentheses indicate the number of labeled samples. Boldfaced numbers indicate the best performing results. ``3s'' means three-stream fusion.}
\label{tab:combined_results}
\fontsize{8}{9.6}\selectfont 
\setlength{\tabcolsep}{2pt} 
\resizebox{\columnwidth}{!}{%
\begin{tabular}{llcccccccccccccccc}
\toprule
\multirow{3}{*}{Method} & \multirow{3}{*}{Stream} & \multicolumn{6}{c}{NTU-60 (\%)} & \multicolumn{2}{c}{NTU-120 (\%)} & \multicolumn{6}{c}{PKU-MMD (\%)} \\ 
\cmidrule(lr){3-8} \cmidrule(lr){9-10} \cmidrule(lr){11-16}
 & & \multicolumn{3}{c}{xsub} & \multicolumn{3}{c}{xview} & xsub & xset & \multicolumn{3}{c}{Part I} & \multicolumn{3}{c}{Part II} \\ 
\cmidrule(lr){3-5} \cmidrule(lr){6-8} \cmidrule(lr){9-9} \cmidrule(lr){10-10} \cmidrule(lr){11-13} \cmidrule(lr){14-16}
 & & 1\% & 10\% & Linear & 1\% & 10\% & Linear & Linear & Linear & 1\% & 10\% & Linear & 1\% & 10\% & Linear \\ 
 & & (400) & (4009) & (40091) & (376) & (3764) & (37646) & Linear & Linear & (188) & (1884) & (18840) & (53) & (533) & (5339) \\ 
\midrule
SkeletonCLR & Joint & 25.7 & 64.6 & 68.3 & 35.4 & 71.4 & 76.4 & 56.8 & 55.9 & 29.9 & 71.2 & 80.9 & 5.0 & 16.4 & 25.9 \\
Ours &  & \textbf{41.0} & \textbf{68.2} & \textbf{71.0} & \textbf{44.9} & \textbf{74.5} & \textbf{76.8} & \textbf{57.5} & \textbf{58.8} & \textbf{53.6} & \textbf{76.9} & \textbf{81.5} & \textbf{7.9} & \textbf{20.3} & \textbf{34.5} \\ 
\midrule
SkeletonCLR & Motion & 17.3 & 62.2 & 53.3 & 27.7 & 68.2 & 50.8 & 39.6 & 40.2 & 18.5 & 71.4 & 63.4 & 4.7 & 7.3 & 15.7 \\
Ours &  & \textbf{37.9} & \textbf{66.3} & \textbf{53.9} & \textbf{40.6} & \textbf{71.6} & \textbf{57.2} & \textbf{44.0} & \textbf{43.7} & \textbf{37.5} & \textbf{73.7} & \textbf{65.2} & \textbf{6.4} & \textbf{15.3} & \textbf{24.4} \\ 
\midrule
SkeletonCLR & Bone & 10.0 & 61.0 & \textbf{69.4} & 30.0 & 65.1 & \textbf{67.4} & 48.4 & \textbf{52.0} & 20.4 & 73.9 & 72.6 & 4.7 & 4.4 & 11.5 \\
Ours &  & \textbf{34.0} & \textbf{68.8} & 61.6 & \textbf{35.8} & \textbf{73.6} & 60.5 & \textbf{52.6} & 49.6 & \textbf{43.0} & \textbf{78.0} & \textbf{76.9} & \textbf{6.0} & \textbf{15.7} & \textbf{24.7} \\ 
\midrule
3s-SkeletonCLR & Fusion & 30.6 & 72.9 & 75.0 & 43.7 & 78.7 & 79.8 & 60.7 & 62.6 & 35.5 & 82.1 & 85.3 & 5.0 & 16.9 & 23.7 \\
Ours &  & \textbf{49.4} & \textbf{76.4} & \textbf{75.2} & \textbf{52.4} & \textbf{81.2} & \textbf{80.2} & \textbf{62.9} & \textbf{63.7} & \textbf{55.3} & \textbf{83.2} & \textbf{86.1} & \textbf{7.6} & \textbf{26.6} & \textbf{39.2} \\ 
\bottomrule
\end{tabular}
}
\vspace{-3mm}
\end{table}

Table~\ref{tab:combined_results} shows the results of our method compared to SkeletonCLR \cite{li20213dhumanactionrepresentation} baseline (further ablation results are provided in Appendix). Under linear evaluation, our method consistently outperforms the baseline on most of results, across all datasets and input streams.  The gains are particularly pronounced for the motion stream (+3.4\% on average), indicating the benefit of explicitly modeling motion variability. Moreover, for multi-stream fusion, our approach outperforms 3s-SkeletonCLR, especially on the more challenging NTU-120 dataset. Overall, these results support our hypothesis that incorporating variational learning leads to more discriminative representations, resulting in improved downstream classification accuracy.





In semi-supervised settings, when only 1\% of labeled samples are available, our method yields substantial accuracy improvements over 3s-SkeletonCLR on both NTU-60 and PKU-MMD. This indicates that the feature representation provided by our variational self-supervised method is able to extract essential characteristics, even under limited examples (i.e.  through the weak supervision signal). Furthermore, as the labeled ratio increases to 10\%, the margin of the accuracy achieved from both methods is decreasing (as the sample sizes grow). However, our approach continues to outperform SkeletonCLR consistently. Importantly, the improvement on PKU\_MMD Part II remains substantial, suggesting that the benefit of our method is not limited to extremely low-data regimes but also extends to moderately supervised conditions. 


\textbf{Latent Features and Attention Analysis}
\begin{figure}[hbt]
\centering
\fontsize{8}{9.6}\selectfont
\begin{subfigure}[b]{0.31\linewidth}
    \centering \textbf{Linear}
\end{subfigure}
\begin{subfigure}[b]{0.31\linewidth}
    \centering \textbf{Semi-10\%}
\end{subfigure}
\begin{subfigure}[b]{0.31\linewidth}
    \centering \textbf{Semi-1\%}
\end{subfigure}

\medskip 

\rotatebox{90}{\quad \textbf{SkeletonCLR}} \hspace{0.5em}
\begin{subfigure}[t]{0.29\linewidth}
    \centering
    \includegraphics[width=\linewidth, trim=0 0 0 6, clip]{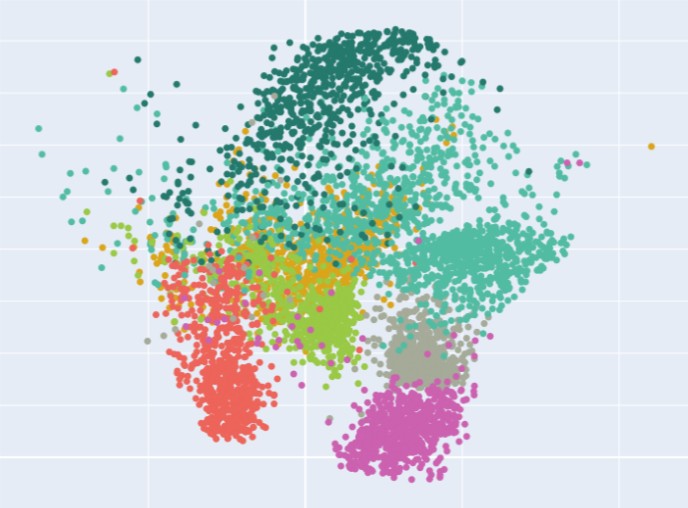}
\end{subfigure}
\begin{subfigure}[t]{0.29\linewidth}
    \centering
    \includegraphics[width=\linewidth, clip]{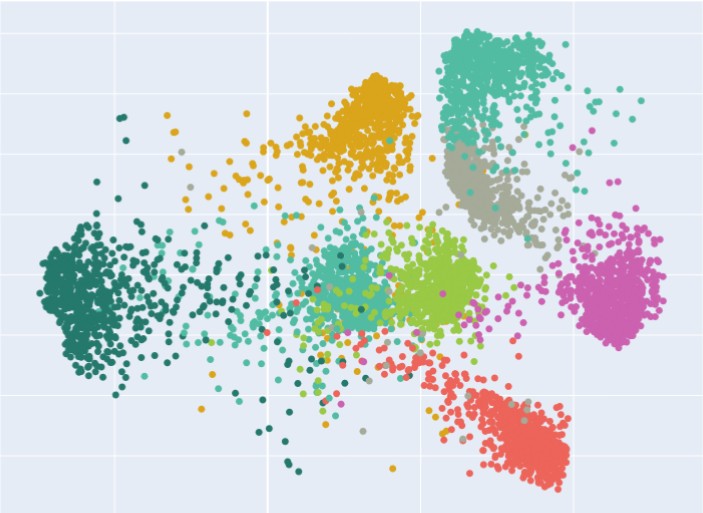}
\end{subfigure}
\begin{subfigure}[t]{0.29\linewidth}
    \centering
    \includegraphics[width=\linewidth, trim=0 0 0 6, clip]{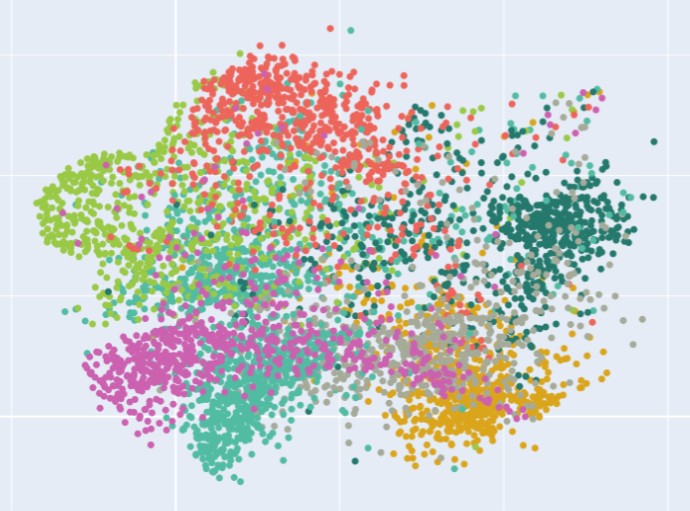}
\end{subfigure}

\vspace{0.5em}

\rotatebox{90}{\qquad \quad \textbf{Ours}} \hspace{0.5em}
\begin{subfigure}[t]{0.29\linewidth}
    \centering
    \includegraphics[width=\linewidth]{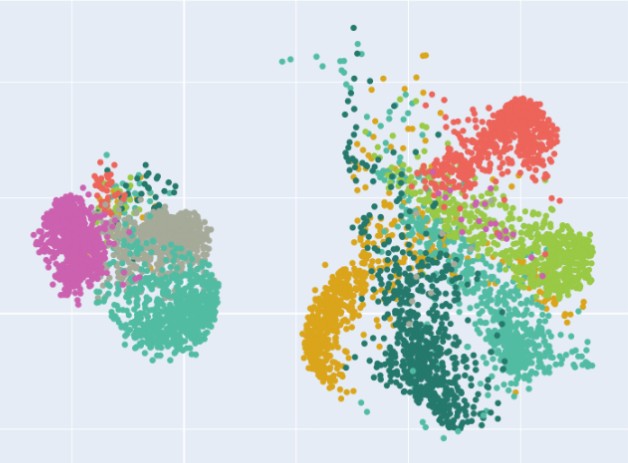}
\end{subfigure}
\begin{subfigure}[t]{0.29\linewidth}
    \centering
    \includegraphics[width=\linewidth]{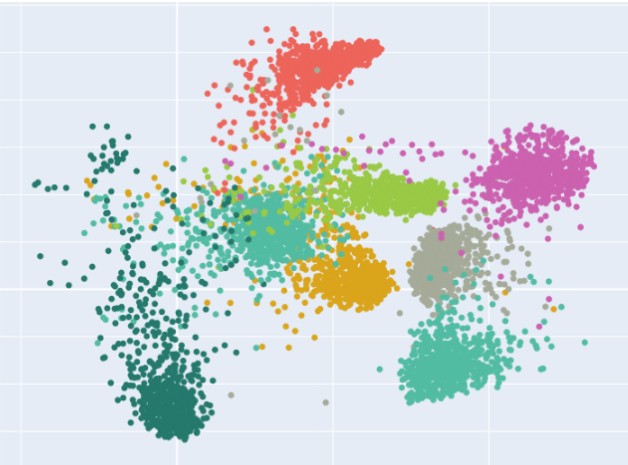}
\end{subfigure}
\begin{subfigure}[t]{0.29\linewidth}
    \centering
    \includegraphics[width=\linewidth]{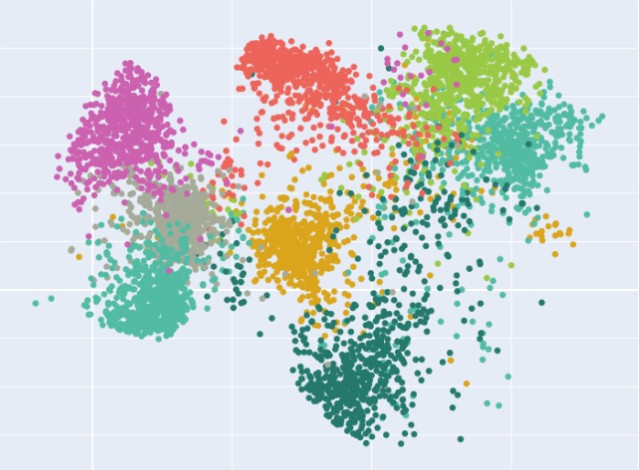}
\end{subfigure}

\caption{Qualitative comparison of the embedding distribution between SkeletonCLR and the proposed method under different supervision settings for NTU-60 dataset.}
\label{fig:qualitative}
\vspace{-3mm}
\end{figure}


We use UMAP embedding~\cite{mcinnes2020umapuniformmanifoldapproximation} with fixed hyperparameters to visualize the embedding distributions with different supervision samples (Fig.~\ref{fig:qualitative}). In the linear evaluation of NTU-60 dataset, with self-supervised pre-training, our method  substantially yields more separable clusters than SkeletonCLR. In particular, the patterns show isolations of the features from two-person actions (left) to one-person actions (right). 

In semi-supervised settings, where supervision signals are reduced, our approach consistently forms tighter, easier to separate clusters  (as opposed to SkeletonCLR). With 1\% labels, we can observe the formation of dense clusters, more distanced to each other when compared to those yielded by the SkeletonCLR approach. Similarly, at 10\% labels the same behavior is observed. This is explained by the incorporation of prior modeling (i.e. variational learning) that  enforces normal distribution as prior, combined with contrastive learning that improves the separability of the feature clusters.  




\begin{figure}[!hbt]
\centering
\fontsize{7.5}{9}\selectfont

\textbf{Action: Eating Meal (Linear)} \\
\vspace{0.3em}
\begin{minipage}{0.04\linewidth}
    \rotatebox{90}{\textbf{SkeletonCLR}}
\end{minipage}
\begin{minipage}{0.94\linewidth}
    \includegraphics[width=\linewidth]{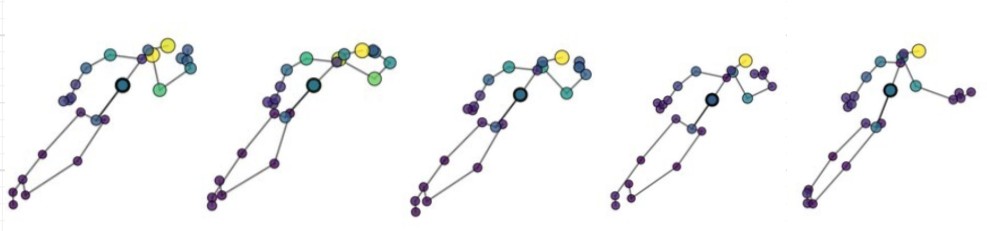}
\end{minipage}

\vspace{0.1em}

\begin{minipage}{0.04\linewidth}
    \rotatebox{90}{\quad \textbf{Ours}}
\end{minipage}
\begin{minipage}{0.94\linewidth}
    \includegraphics[width=\linewidth]{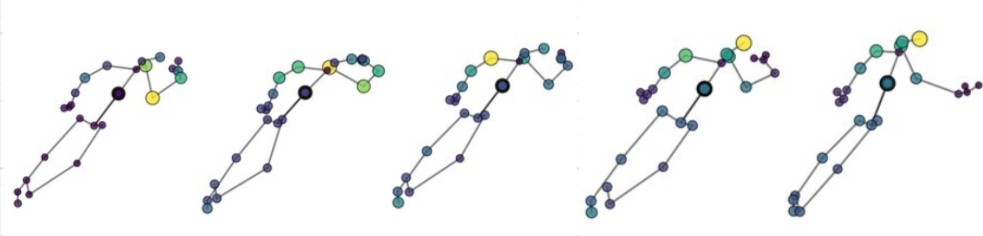}
\end{minipage}

\vspace{0.2em}
\hspace{0.04\linewidth} 
\begin{minipage}{0.94\linewidth}
    \centering
    \begin{tabularx}{\linewidth}{@{}YYYYY@{}}
        $t=10$ & $t=25$ & $t=35$ & $t=45$ & $t=50$
    \end{tabularx}
\end{minipage}
    
\vspace{0.8em} 

\textbf{Action: Taking off Jacket (Semi-1\%)} \\
\vspace{0.3em}
\begin{minipage}{0.04\linewidth}
    \rotatebox{90}{\textbf{SkeletonCLR}}
\end{minipage}
\begin{minipage}{0.94\linewidth}
    \includegraphics[width=\linewidth]{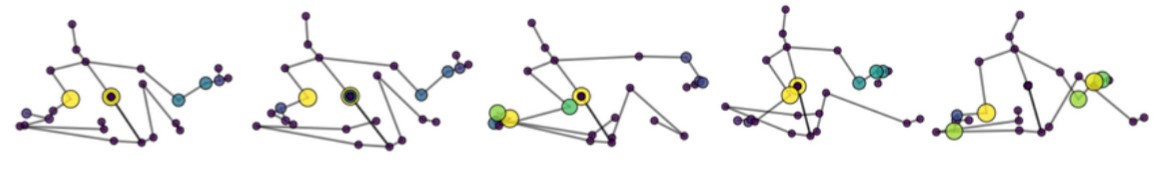}
\end{minipage}

\vspace{0.2em}

\begin{minipage}{0.04\linewidth}
    \rotatebox{90}{\quad \textbf{Ours}}
\end{minipage}
\begin{minipage}{0.94\linewidth}
    \includegraphics[width=\linewidth]{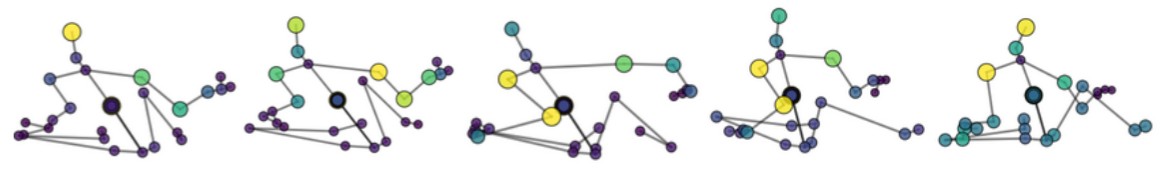}
\end{minipage}

\vspace{0.2em}
\hspace{0.04\linewidth} 
\begin{minipage}{0.94\linewidth}
    \centering
    \begin{tabularx}{\linewidth}{@{}YYYYY@{}}
        $t=10$ & $t=25$ & $t=35$ & $t=45$ & $t=50$
    \end{tabularx}
\end{minipage}

\caption{Grad-CAM visualizations of joints importance of SkeletonCLR with our method for different actions and supervisions. Brighter-colored joints indicate higher relevance.}
\label{fig:attention}
\vspace{-3mm}
\end{figure}

Fig.~\ref{fig:attention} shows Grad-CAM attention \cite{Selvaraju_2019} of SkeletonCLR and our method on linear and semi-supervised settings applied to joint inputs of samples from NTU-60. For the "Eating Meal" action, SkeletonCLR primarily attends  the head area and fails to capture the characteristic hand-to-mouth motion. In contrast, our method consistently emphasizes the relevant body regions (ankle and arm joints) involved in the action, enabling more accurate modeling of the motion dynamics.

On the second example ("Taking of Jacket", 1\%), we can observe more pronounced differences. Here, SkeletonCLR focuses are largely stagnant (mostly on a specific joint and notably missing the left ankle), while relevant movements are occurring at the other joints (in particular the body joints starting from the 35th frame onward). In contrast, our method is able to focus on such joints, even under limited supervision. 


\subsection{Comparison with state-of-the-art}

In this section, we provide the comparison of our method with state-of-the-art techniques using the respective protocols and evaluation metrics, produced based on either their original implementations or reported results.

\textbf{Linear and Finetuning Results}

\begin{table}[!ht]
\centering
\caption{Comparison of linear evaluation results on NTU-60, NTU-120, and PKU-MMD datasets. \textbf{Bold}, \textcolor{red}{red}, and \textcolor{blue}{blue} indicate the first, second, and third best results, respectively.}
\label{tab:sota_linear}
\fontsize{7.5}{9}\selectfont 
\setlength{\tabcolsep}{3pt} 
\resizebox{\columnwidth}{!}{%
\begin{tabular}{llcccccccc}
\toprule
\multirow{2}{*}{Method} & \multirow{2}{*}{Backbone} & \multirow{2}{*}{Params} & \multicolumn{2}{c}{NTU-60 (\%)} & \multicolumn{2}{c}{NTU-120 (\%)} & \multicolumn{2}{c}{PKU-MMD (\%)} \\
\cmidrule(lr){4-5} \cmidrule(lr){6-7} \cmidrule(lr){8-9}
 & & & xsub & xview & xsub & xset & Part I & Part II \\
\midrule
LongT GAN \cite{zheng2018unsupervised} & GRU & 40.2M & 39.1 & 48.1 & 35.6 & 39.7 & 67.7 & 26.0 \\
MS2L \cite{Lin_2020} & GRU & 11.4M & 52.6 & -- & -- & -- & 64.9 & 27.6 \\
AS-CAL \cite{rao2021augmentedskeletonbasedcontrastive} & LSTM & 0.4M & 58.5 & 64.8 & 48.6 & 49.2 & -- & -- \\
P\&C \cite{su2020predict} & GRU & 57.5M & 50.7 & 76.3 & 42.7 & 41.7 & -- & -- \\
SeBiReNet \cite{nie2021unsupervised3dhumanpose} & GRU & 0.3M & -- & 79.7 & -- & -- & -- & -- \\
TS-Colorization \cite{yang2021skeletoncloudcolorizationunsupervised} & DGCNN & -- & 71.6 & 79.9 & -- & -- & -- & -- \\
ISC \cite{thoker2021skeletoncontrastive3dactionrepresentation} & GRU+GCN & 106.0M & -- & -- & \textbf{67.9} & \textcolor{blue}{67.1} & \textcolor{blue}{80.9} & \textcolor{red}{36.0} \\
3s-CrosSCLR \cite{li20213dhumanactionrepresentation} & GCN & 5.7M & \textbf{77.8} & \textcolor{red}{83.4} & \textbf{67.9} & \textcolor{red}{66.7} & \textcolor{red}{84.9} & \textcolor{blue}{21.2} \\
GL-Transformer \cite{kim2022globallocalmotiontransformerunsupervised}& Transformer & 214.4M & \textcolor{red}{76.3} & \textbf{83.8} & \textcolor{red}{66.0} & \textbf{68.7} & -- & -- \\
\midrule
\textbf{Ours} & GCN & 1.9M & \textcolor{blue}{75.2} & \textcolor{blue}{80.2} & \textcolor{blue}{62.9} & 63.7 & \textbf{86.1} & \textbf{39.2} \\
\bottomrule
\end{tabular}
}
\vspace{-3mm}
\end{table}

Table \ref{tab:sota_linear} shows the results for linear evaluation on the NTU-60 dataset, where our method outperforms earlier self-supervised approaches such as LongT GAN, MS2L, AS-CAL, P\&C, SeBiReNet and TS-Colorization while remains comparable to those of CrosSCLR and GL-Transformer under both xsub and xviews protocols. It is worth noting that our competitive results are achieved with a lightweight ST-GCN backbone (with less than 2M parameters) and a simpler training pipeline vs 3s-CrosSCLR that implements more elaborate mining strategy and GL-Transformer which is based on transformer architecture (that is far more complex than a normal ST-GCN backbone, with more than 214M parameters).



On the NTU-120 dataset, our method achieves competitive performance compared to recent self-supervised baselines (such as CrosSCLR and GL-Transformer), while still outperforming several earlier approaches. This shows that the proposed method is able to generalize to larger-scale and more diverse action categories. On PKU-MMD dataset, our approach achieves the best performance on Part I (86.02\%), improving upon 3s-CrosSCLR by +1.1\%, and offers a substantial gain on the challenging Part II, where accuracy increases from 21.2\% to 39.2\% (+18.0\%), thus highlighting its robustness.

\begin{table}[!hbt]
\centering
\caption{Finetuning evaluation results on NTU-60 and NTU-120 datasets. Best scores are in bold.}
\label{tab:sota_finetune}
\fontsize{8}{9.6}\selectfont
\setlength{\tabcolsep}{10pt} 
\begin{tabular}{lcccc}
\toprule
\multirow{2}{*}{Method} & \multicolumn{2}{c}{NTU-60 (\%)} & \multicolumn{2}{c}{NTU-120 (\%)} \\ 
\cmidrule(lr){2-3} \cmidrule(lr){4-5}
 & xsub & xview & xsub & xset \\ 
\midrule
3s-ST-GCN \cite{yan2018spatialtemporalgraphconvolutional} & 85.2 & 91.4 & 77.2 & 77.1 \\
3s-CrosSCLR \cite{li20213dhumanactionrepresentation} & 86.2 & 92.5 & \textbf{80.5} & 80.4 \\
Ours            & \textbf{86.6} & \textbf{92.9} & 79.8 & \textbf{81.4} \\ 
\bottomrule
\end{tabular}
\vspace{-3mm}
\end{table}

Table~\ref{tab:sota_finetune} reports fine-tuning results on NTU-60 and NTU-120. For fair comparisons, the ST-GCN used in the methods of this table all have the same structure and parameters. In this fully supervised setting, we can also see the classification results of our methods are in majority higher than the others. Compared to 3s-CrosSCLR, our approach achieves higher accuracy on NTU-60 under both xsub and xview settings, while remaining competitive on NTU-120. 

These results demonstrate that the proposed method is capable to effectively leverage all labeled samples and benefits from the strong representations learned during pretext training.

\textbf{Semi-supervised with Fewer Labels}

\begin{table}[!hbt]
\centering
\caption{Semi-supervised evaluation on the NTU RGB+D 60 and PKU-MMD datasets. Values in parentheses indicate the number of labeled samples. \textbf{Bold}, \textcolor{red}{red}, and \textcolor{blue}{blue} fonts denote the 1st, 2nd, and 3rd highest accuracies, respectively.}
\label{tab:sota_semi}
\fontsize{8}{9.6}\selectfont 
\setlength{\tabcolsep}{3pt} 
\resizebox{\columnwidth}{!}{%
\begin{tabular}{lcccccccc}
\toprule
\multirow{3}{*}{Method} & \multicolumn{4}{c}{NTU-60 (\%)} & \multicolumn{4}{c}{PKU-MMD (\%)} \\ 
\cmidrule(lr){2-5} \cmidrule(lr){6-9}
 & \multicolumn{2}{c}{xsub} & \multicolumn{2}{c}{xview} & \multicolumn{2}{c}{Part I} & \multicolumn{2}{c}{Part II} \\ 
\cmidrule(lr){2-3} \cmidrule(lr){4-5} \cmidrule(lr){6-7} \cmidrule(lr){8-9}
 & 1\%  & 10\% & 1\% & 10\% & 1\% & 10\% & 1\% & 10\% \\ 
 & (400)  & (4009) & (376)& (3764) & (188) & (1884) & (53) & (533) \\
\midrule
LongT GAN \cite{zheng2018unsupervised} & 35.2 & 62.0 & --   & --   & 35.8 & 69.5 & \textcolor{red}{12.4} & 25.7 \\
MS2L \cite{Lin_2020} & 33.1 & 65.2 & --   & --   & 36.4 & 70.3 & \textbf{13.0} & \textcolor{blue}{26.1} \\
ISC \cite{thoker2021skeletoncontrastive3dactionrepresentation} & 35.7 & 65.9 & 38.1 & 72.5 & \textcolor{blue}{37.7} & 72.1 & -- & -- \\
TS-Colorization \cite{yang2021skeletoncloudcolorizationunsupervised} & \textcolor{blue}{42.9} & 66.1 & \textcolor{blue}{46.3} & 73.3 & -- & -- & -- & -- \\
3s-SkeletonCLR \cite{li20213dhumanactionrepresentation} & 30.6 & \textcolor{blue}{72.9} & 43.7 & \textcolor{red}{78.7} & 35.5 & \textcolor{blue}{82.1} & 5.0 & 16.9 \\
3s-CrosSCLR \cite{li20213dhumanactionrepresentation} & \textbf{51.1} & \textcolor{red}{74.4} & \textcolor{red}{50.0} & \textcolor{blue}{77.8} & \textcolor{red}{49.7} & \textcolor{red}{82.9} & \textcolor{blue}{10.2} & \textbf{28.6} \\
\midrule
\textbf{Ours}   & \textcolor{red}{49.4} & \textbf{76.4} & \textbf{52.4} & \textbf{81.2} & \textbf{55.3} & \textbf{83.2} & 7.6 & \textcolor{red}{26.6} \\ 
\bottomrule
\end{tabular}
}
\vspace{-3mm}
\end{table}

Table \ref{tab:sota_semi} presents semi-supervised evaluation results with 1\% and 10\% labeled data, comparing our method with existing self-supervised approaches on NTU-60 and PKU-MMD. In overall, our method consistently outperforms previous approaches across different sample sizes.

In 1\% labeled data setting, our method produces high accuracy results on both datasets. On NTU-60, we obtain 49.4\% (xsub) and 52.4\% (xview) accuracy rates, which substantially outperforms most of the earlier approaches. Compared to more recent methods such as 3s-SkeleonCLR and 3s-CrosSCLR, our results achieve the best xview accuracy, closely matching the best xsub result. On PKU-MMD, our method achieves the highest accuracy on Part I (54.5\%), surpassing 3s-CrosSCLR by +4.8\%. With 10\% labeled data, our approach continues to deliver consistent gains. On NTU-60, it achieves 76.0\% (xsub) and 80.2\% (xview), outperforming earlier baselines and matching or surpassing recent methods such as 3s-SkeletonCLR and 3s-CrosSCLR. Finally, on PKU-MMD, our method also achieves the best performance on Part I (83.2\%), while remaining competitive on the more challenging Part II.




\begin{figure}[!hbt]
\centering
\fontsize{7.5}{9}\selectfont

\textbf{UMAP Visualization (PKU-MMD Part I)} \\
\vspace{0.3em}

\hspace{0.04\linewidth}
\begin{minipage}{0.31\linewidth}\centering \textbf{Linear}\end{minipage}
\begin{minipage}{0.31\linewidth}\centering \textbf{Semi-10\%}\end{minipage}
\begin{minipage}{0.31\linewidth}\centering \textbf{Semi-1\%}\end{minipage} \\
\vspace{0.2em}

\begin{minipage}{0.04\linewidth}
    \rotatebox{90}{\textbf{CrosSCLR}}
\end{minipage}
\begin{minipage}{0.94\linewidth}
    \includegraphics[width=0.32\linewidth]{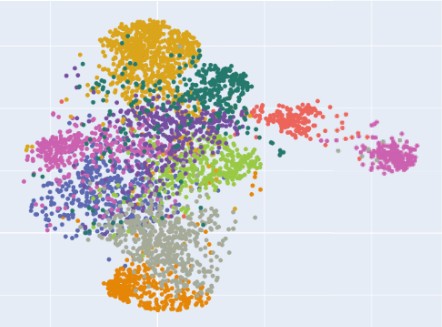}\hfill
    \includegraphics[width=0.32\linewidth]{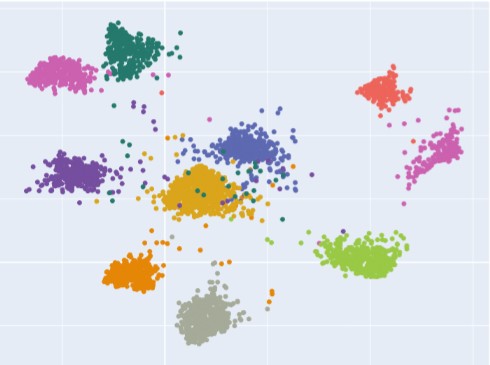}\hfill
    \includegraphics[width=0.32\linewidth]{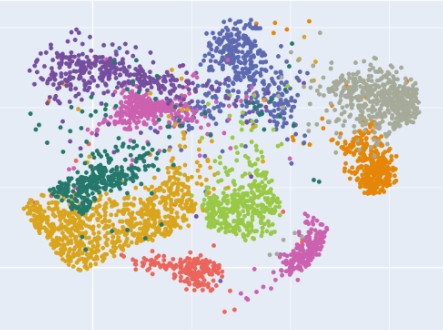}
\end{minipage}

\vspace{0.3em}

\begin{minipage}{0.04\linewidth}
    \rotatebox{90}{\quad \textbf{Ours}}
\end{minipage}
\begin{minipage}{0.94\linewidth}
    \includegraphics[width=0.32\linewidth]{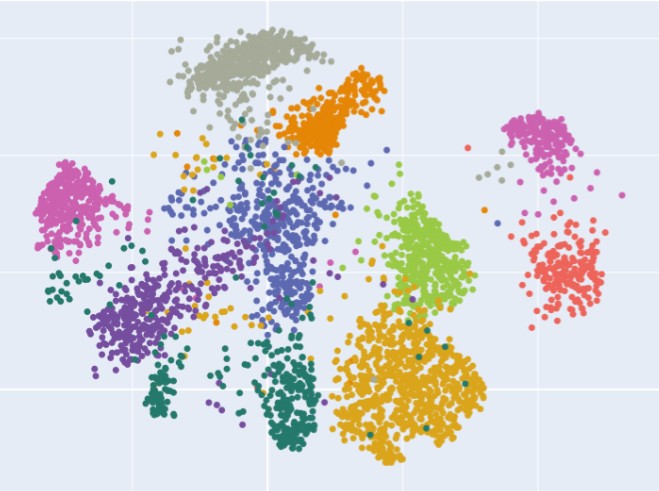}\hfill
    \includegraphics[width=0.32\linewidth]{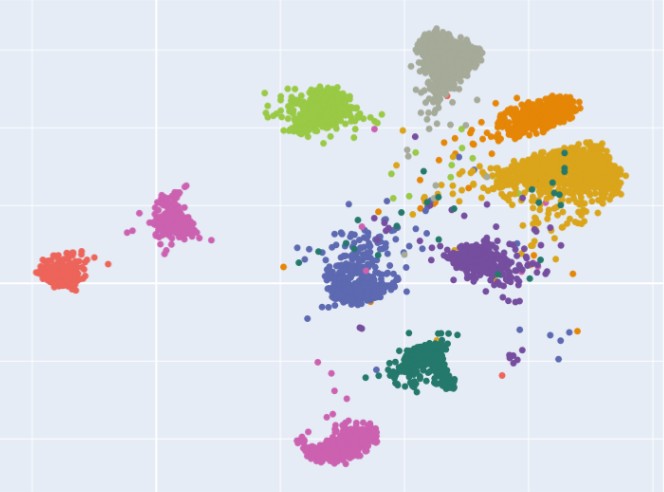}\hfill
    \includegraphics[width=0.32\linewidth]{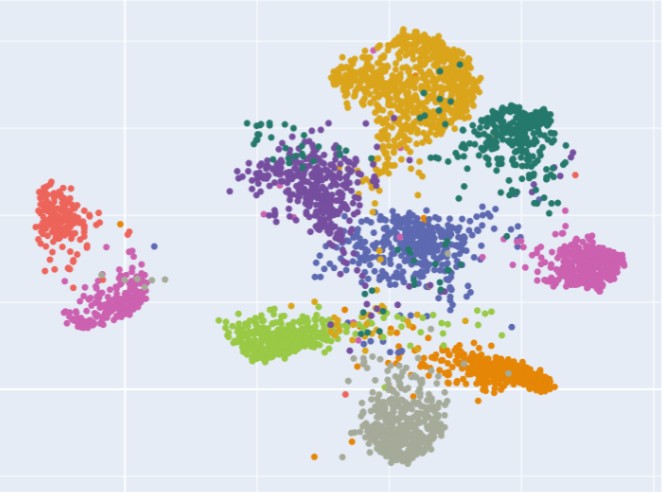}
\end{minipage}

\vspace{1.2em} 

\textbf{Attention Focus (Semi-1\% Setting)} \\
\vspace{0.3em}

\begin{minipage}{0.04\linewidth}
    \rotatebox{90}{\textbf{CrosSCLR}}
\end{minipage}
\begin{minipage}{0.94\linewidth}
    \includegraphics[width=\linewidth]{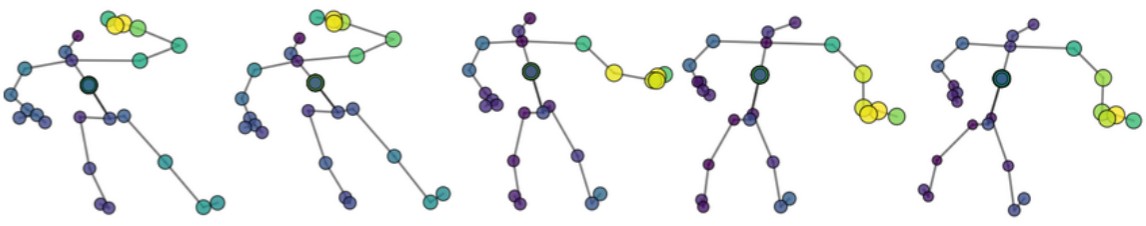}
\end{minipage}

\vspace{0.2em}

\begin{minipage}{0.04\linewidth}
    \rotatebox{90}{\quad \textbf{Ours}}
\end{minipage}
\begin{minipage}{0.94\linewidth}
    \includegraphics[width=\linewidth]{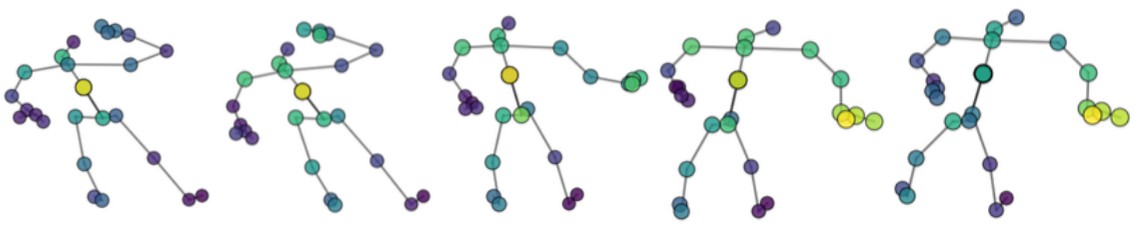}
\end{minipage}

\vspace{0.2em}
\hspace{0.04\linewidth}
\begin{minipage}{0.94\linewidth}
    \centering
    \mbox{\makebox[0.2\linewidth]{$t=10$}\makebox[0.2\linewidth]{$t=25$}\makebox[0.2\linewidth]{$t=35$}\makebox[0.2\linewidth]{$t=45$}\makebox[0.2\linewidth]{$t=50$}}
\end{minipage}

\caption{Qualitative results on PKU-MMD Part I. \textbf{Top:} UMAP visualizations across supervision levels of CrosSCLR and ours. \textbf{Bottom:} Grad-CAM attention focus over time for the 1\% labeled setting. Brighter joints signify higher relevance.}
\label{fig:sota_qualitative}
\vspace{-3mm}
\end{figure}

Fig.~\ref{fig:sota_qualitative} illustrates the learned representations  PKU-MMD Part I under different supervision levels. The results demonstrate that our method yields more structured and separable embeddings than CrosSCLR (which also uses complex mining strategies, apart of a traditional contrastive learning). This embedding characteristics can be an advantage, particularly on pre-trained representations in low-level 1\% labeled setting.



Finally, the attention analysis (bottom) demonstrates the ability of our method to focus on specific joints movements (such as right arms), but also to include other relevant joints, such as shoulder joints (that are not focused by CrosSCLR). This  helps obtaining more reliable downstream predictions, explaining the competitive overall performance of our method.


\section{Conclusion and perspectives}

In this work, we have addressed the limitations of current self-supervised skeleton-based action recognition, where existing approaches remain insensitive to the intrinsic structure and dynamics of human motion. To do this, we propose a variational contrastive framework that combines probabilistic latent modeling with contrastive self-supervised learning to obtain structured and discriminative representations. Our method jointly optimizes variational and contrastive objectives within a unified training pipeline, supporting both pretraining and evaluation, particularly under limited labeled data.


The ablation studies performed shows that our method consistently outperforms the baseline results, particularly in low-label regimes. Evaluations on NTU-60, NTU-120, and PKU-MMD datasets further highlights the competitive results of our approach across evaluation protocols. Analysis of the learned representations with limited samples illustrates the relevant motion features generated by our approach compared to existing approaches, a factor that explains its strong quantitative results. Further inspection of model attention confirms that the focus aligns well with action-relevant joints. Future work will investigate multi-modal disentanglement alongside more expressive probabilistic formulations that explicitly model the geometry of contrastive embeddings, further improving representation robustness and downstream recognition performance.

%
%
%
\bibliographystyle{IEEEtranBST2/IEEEtranS.bst}
\bibliography{IEEEtranBST2/IEEEexample,IEEEtranBST2/IEEEfull}

\end{document}